\documentclass[letterpaper,10pt,conference]{ieeeconf}
\IEEEoverridecommandlockouts
\usepackage{booktabs,graphicx,amsmath,amssymb,xcolor}

\title{\LARGE \bf Beyond End-Task Success:\\How to Audit Visual Experience Retrieval in Robotics}

\author{Eshika Pathak$^{1,2}$ and Leela Krishna$^{2}$%
\thanks{Accepted to the IROS 2026 Workshop on Embodied Neuro-Symbolic AI for Reliable and Safe Robotics (ReS AI).}%
\thanks{$^{1}$University of Illinois Urbana-Champaign. $^{2}$Centific. Work by E.~Pathak was done while at Centific. Contact: \texttt{epathak2@illinois.edu}}%
}

\begin{document}
\maketitle
\begin{abstract}
Robots that store past experiences must select which one to reuse in a new scene. Most systems select by visual similarity, and most evaluations report only the success of the selected experience. That number does not show whether the selection was good: a rule can score well by repeatedly using one broadly transferable experience, or poorly because its preferred experience is weak. This matters because robots increasingly adapt by reuse rather than retraining: every such adaptation passes through this selection step, and a score that describes the library rather than the rule misleads what the field builds next. We contribute an audit methodology: execute every stored experience in every query scene, yielding the complete table of transfer outcomes, over two manipulation tasks, three reuse mechanisms, and libraries of $K=3$, $10$, and $50$. The complete table is what makes the confound measurable: every alternative's outcome is known, so a score can be traced to per-scene selection or to library quality. The audited rules select by nearest-neighbor distance in five visual embeddings, from raw pixels to CLIP. We see four results. (1) One fixed experience, chosen with hindsight, captures 30--58\% of the gap between random selection and an oracle; per-scene selection competes for the remaining 0.07--0.15 in success rate. (2) At $K\ge10$, visual rules concentrate on one experience 1.5--3 times more than the oracle does, and their scores then follow that experience's quality. (3) Wherever a rule differs significantly from a shuffle that keeps its selection rates but pairs them with scenes at random, the rule is worse, for every learned image policy. (4) Visual distance predicts well whether a given pair will succeed (AUROC up to 0.96), yet ranks the candidates within one scene no better than chance for four of five embeddings at $K=50$ (AUROC 0.45--0.52). Exhaustive execution is usually infeasible, so the audit reduces to two cheap reports any study can give: the distribution of selected experiences, and the success of the best single experience in hindsight.
\end{abstract}

\section{Introduction}
Robots increasingly store demonstrations, trajectories, and policies for reuse. Reuse is becoming how robots adapt in deployment: a retrieved demonstration seeds few-shot imitation, a similar past episode conditions a policy, a stored skill is executed directly. Each of these passes through one selection step. Given a new scene, the robot must select which stored experience to use. A common rule is nearest-neighbor retrieval: embed the current scene, embed each stored scene, and select the experience whose stored scene is closest~\cite{pari2022vinn,dipalo2024dinobot,kwon2025rtcache}. The rule assumes that visually similar scenes call for the same experience.

We study how to evaluate this selection. For a query scene $q$ and a stored experience $e_i$, define the transfer probability
\begin{equation}
T(i,q)=P(\mathrm{success}\mid e_i,q).
\label{eq:T}
\end{equation}
An ideal retriever selects $\arg\max_i T(i,q)$. A similarity rule instead selects $\arg\min_i d(f(q),f(e_i))$, where $f$ is an embedding and $d$ a distance. The audit asks how well the second quantity approximates the first.

Standard evaluations execute only the selected experience and report its success. This number mixes two things: the quality of the decision and the quality of the library. A rule can score well by always selecting one strong experience. A rule can score poorly because it keeps selecting one weak experience. In both cases the score describes an experience, not the rule.

We therefore ask a methodological question: \emph{how can we audit whether a robot retriever actually chose well?} Our answer is to measure the alternatives. In simulation, we execute every stored experience in every query scene. This yields the complete \emph{transfer matrix} of outcomes, which no ordinary deployment can observe. We also vary scene \emph{appearance} (lighting, color, texture, camera pose) independently of scene \emph{layout} (object poses, articulation, obstacles). The physics is identical across appearance changes, so appearance effects and layout effects can be separated.

The audit has one central finding. Predicting transfer and selecting an experience are different problems. Visual distance predicts well whether a given experience-scene pair will succeed. It ranks poorly which of the $K$ candidates to use for a given scene. The reason is that most of the outcome variance is shared within a scene: a hard scene is hard for every candidate. A score can detect hard scenes and still not separate the candidates. This one distinction explains three observations that would otherwise conflict: visual rules concentrate on few experiences, equalizing library quality restores positive discrimination, and high pooled prediction accuracy coexists with negative selection value.

The contribution is methodological. We propose no new retrieval method, and we claim no general verdict on visual similarity. We propose a way to measure whether a retriever chose well, and we show that this measurement changes the conclusions an ordinary evaluation would draw, including several of our own.

\textbf{Contributions.}
(1) An audit protocol that scores retrieval against the complete transfer matrix, not only the selected trajectory.
(2) Direct measurements separating pooled transfer prediction from within-query ranking, showing that the first does not imply the second.
(3) Controls that isolate the effect of library composition: an oracle concentration reference, a best-fixed reference, frequency-matched shuffles, random sub-libraries, and quality-equalized libraries.
(4) A directional result for learned image policies: whenever visual selection differs significantly from the frequency-matched baseline, it is worse.
(5) Two reporting diagnostics for ordinary studies: the selection distribution and the best-fixed reference.

\section{Audit protocol}
\label{sec:setup}
\textbf{Tasks and scenes.} We use Stack and Door in robosuite~\cite{zhu2020robosuite} with a Panda arm under pose control. A scripted expert records each stored experience in its own base scene. Each query set contains 40 physical states. The states span four layout levels: object displacement up to $\pm12$\,cm, relative displacement or articulation, and obstacles. Each state is rendered under five appearance draws. Appearance draws change lighting, color, texture, and small camera rotations. They do not change physics: the simulator state hash is identical across the five renders of each layout.

\textbf{Reuse mechanisms.} We evaluate three ways to reuse an experience. \emph{Replay} retargets the recorded end-effector path to the current object poses and executes it open-loop~\cite{mandlekar2023mimicgen}. Replay never reads the camera, so appearance cannot affect it. This is confirmed in the data: 0 of 600 labelled cases change across renders. \emph{Policy} trains one behavior-cloned image policy per experience (BC-RNN~\cite{mandlekar2021robomimic}). Both appearance and layout affect a Policy execution. \emph{Policy+DR} trains the same policy class with randomized lighting and texture. On Stack, this reduces the share of failures caused by appearance from 26\% to 2\%, and layout-caused failures remain matched pair by pair. On Door, domain randomization also degrades layout performance.

\textbf{Ground truth.} We execute every pair $(e_i,q)$ with five seeds and record the success rate $y_{iq}$. The result is an answer that ordinary deployment cannot see: for each scene, the outcome of the selected experience, of every experience not selected, of random selection, and of oracle selection. The study uses $114{,}500$ seeded executions plus 250 own-base replays. The unit of analysis is the transfer matrix, not the execution count.

\textbf{Selection rules.} We test nearest-neighbor retrieval in five embeddings: raw $32\times32$ pixels, ImageNet ResNet-18, DINOv2~\cite{oquab2024dinov2}, CLIP~\cite{radford2021clip}, and the Policy+DR visual encoder. We also test recency and two structural scores: object-position proximity and path clearance. Finally, a track-record rule ranks experiences by their outcomes on past scenes. Track record uses outcome data that a zero-shot visual rule does not have. We use it only as a reference for how quickly outcome data can find a strong experience.

\textbf{Metrics.} We report three quantities. First, the \emph{normalized advantage}
\begin{equation}
A=\frac{S_{\mathrm{rule}}-S_{\mathrm{random}}}{S_{\mathrm{oracle}}-S_{\mathrm{random}}},
\end{equation}
where $S$ is mean success: $A{=}0$ matches random selection and $A{=}1$ matches the oracle. Second, raw success rates. Third, the \emph{discrimination gap}. To compute it, we generate 2{,}000 shuffled rules. Each shuffled rule selects each experience exactly as often as the real rule, but assigns those selections to scenes at random. The gap is the real rule's success minus the mean shuffled success. A positive gap means the rule's scene assignment adds value beyond its selection frequencies. Intervals are clustered by scene ($n{=}40$). We omit $A$ when its denominator falls below a threshold fixed in advance.

\textbf{Advance predictions.} We wrote down predictions and analysis thresholds before collecting the data that tested them. Where the text says a prediction was made ``in advance'', it refers to this record. Several of our own predictions failed, and we note the failures where they change the interpretation.

\section{How much can selection add?}
\label{sec:headroom}
Selection matters only if different scenes need different experiences. We first measure how much room the benchmark leaves for this.

Define \emph{best-fixed} as the single experience with the highest mean success on the query set, chosen after seeing all outcomes. Best-fixed uses no per-scene information. It is not a deployable method; it is a reference.

Best-fixed is strong (Table~\ref{tab:inv}). It captures 30--58\% of the gap between random and oracle selection in every stable setting; in the advantage metric, $A=0.30$--$0.58$ (Table~\ref{tab:inv}). No tested rule beats it with non-overlapping intervals, and point excesses are at most 0.04. The intervals are wide, so the correct conclusion is that no rule is distinguishable from best-fixed, not that all rules equal it.

The headroom above best-fixed is small (Table~\ref{tab:inv}, headroom column). The oracle exceeds best-fixed by only 0.07--0.15 in success rate, in every stable setting. The headroom does not grow with library size: at $K{=}50$ it is 0.130, below our advance prediction of at least 0.15. The conclusion is direct. In these libraries, the main opportunity is not to match experiences to scenes. It is to find the experiences that transfer broadly. Section~\ref{sec:pairwise} shows the same fact from the variance side.

\begin{table}[t]
\centering\scriptsize\setlength{\tabcolsep}{2.5pt}
\caption{Best-fixed advantage, oracle and random success, headroom (oracle minus best-fixed success), and the oracle's top-pick share (the concentration reference of Sec.~\ref{sec:inherit}).}
\label{tab:inv}
\begin{tabular}{lrrrrrr}
\toprule
set & $K$ & best-fixed $A$ & oracle & random & headroom & oracle modal\\
\midrule
Stack Replay, orig. & 3 & .58 & .80 & .53 & .110 & .45\\
Stack Replay, fresh & 3 & .45 & .88 & .63 & .135 & .38\\
Stack Replay, A & 10 & .45 & .82 & .55 & .150 & .20\\
Stack Replay, B & 10 & .51 & .86 & .62 & .115 & .20\\
Stack Replay & 50 & .51 & .83 & .56 & .130 & .10\\
Door P/+DR, orig. & 3 & .36/.55 & .78/.77 & .64/.61 & .088/.068 & .41/.50\\
Door P/+DR, fresh & 3 & .30/.40 & .81/.82 & .66/.67 & .103/.092 & .47/.50\\
\bottomrule
\end{tabular}
\end{table}

Outcome data finds strong experiences quickly. The track-record rule selects the best or second-best experience in every set, and it locks onto a strong experience after about six scenes. On fresh image-policy scenes it matches or beats every visual rule in the warm-start setting. At $K{=}50$ it still trails best-fixed ($A{=}0.26$ versus $0.51$): estimating 50 candidates from 40 scenes has an exploration cost. Track record is not a fair competitor to visual retrieval, because it sees outcomes. Its role here is to show that a little outcome data identifies the quantity that matters most, the mean success of each experience.

\section{Concentration: the score follows the top pick}
\label{sec:inherit}
Visual rules keep selecting the same experience. We call the experience a rule selects most often its \emph{top pick}. This section measures how concentrated the rules are, and what the top pick does to the score.

Concentration must be compared to the right reference. The oracle itself is concentrated, because several experiences often tie for the best outcome. The oracle's top-pick share is 0.38--0.52 at $K{=}3$ across all settings (Table~\ref{tab:inv} lists the Replay and Door settings), 0.20 at $K{=}10$, and 0.10 at $K{=}50$. The visual rules' top-pick share is 0.50--0.68 in several $K{=}3$ settings, 0.30--0.60 at $K{=}10$, and 0.25--0.30 at $K{=}50$. Against the oracle reference, the excess at $K{=}3$ is small and often absent. At $K\ge10$ it is clear: visual rules are 1.5--3 times as concentrated as the oracle (Fig.~\ref{fig:k}, left).

Concentration is costly when the top pick is weak. Define the \emph{quality margin} of an experience as its mean success minus the library mean. If a score merely follows the top pick, the sign of the advantage $A$ should match the sign of the top pick's margin: a strong top pick gives $A>0$, a weak one gives $A<0$. We tested this across the whole benchmark. A \emph{setting} is one combination of task, reuse mechanism, and library; 13 settings have a denominator large enough to report $A$ (Sec.~\ref{sec:setup}). Seven rules report $A$ in each of them: the five visual embeddings and the two structural scores. That gives $7\times13=91$ cases. The signs match in 69 of the 91. The policy encoder is the clearest example (Table~\ref{tab:pe}). Its advantage is negative when its top pick is weak, positive when its top pick is strong, and negative again at $K{=}50$, where its top pick is the weakest experience in the library. We predicted the signs of the two $K{=}10$ rows in advance, from the top pick's quality alone. The pattern is not a law. Raw pixels at $K{=}10$ are a counterexample: their top pick is weak, yet they select it mainly on the scenes where it works.

\begin{table}[t]
\centering\scriptsize
\caption{The policy encoder on Stack Replay. Its advantage follows the quality of its top pick.}
\label{tab:pe}
\begin{tabular}{llr}
\toprule
set & top pick & advantage\\
\midrule
$K{=}3$, orig. & below library mean & $-0.77$\\
$K{=}3$, fresh & below library mean & $-0.73$\\
$K{=}10$, A & above library mean & $+0.29$\\
$K{=}10$, B & above library mean & $+0.32$\\
$K{=}50$ & weakest in library (mean 0.28) & $-0.57$\\
\bottomrule
\end{tabular}
\end{table}

\begin{figure}[t]
\centering
\begin{minipage}{0.06\columnwidth}
\centering\rotatebox{90}{\scriptsize discrimination gap}
\end{minipage}%
\begin{minipage}{0.91\columnwidth}
\centering
\includegraphics[width=\linewidth,trim=0.65cm 0 0 0,clip]{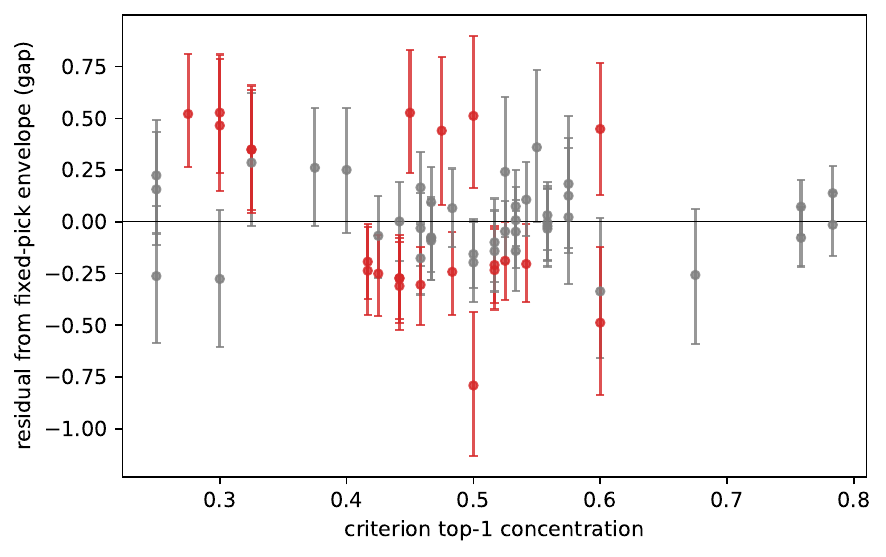}
\end{minipage}
\caption{Discrimination gap versus top-pick share for the visual rules. Error bars show the 95\% envelope of frequency-matched shuffles. Red points lie outside the envelope. The gap shows no clear relation to concentration: concentration measures how often a rule repeats its top pick, not whether its scene assignment helps.}
\label{fig:env}
\end{figure}

Does a rule's scene assignment add anything beyond its frequencies? The discrimination gap answers this. Figure~\ref{fig:env} shows each visual rule's gap with the 95\% envelope of its frequency-matched shuffles. In 42 of 65 cases, the rule lies inside the envelope. This does not prove those rules are equivalent to frequency-matched random assignment. With 40 scenes the envelope is wide, and we have not measured its power to detect small gaps. It shows compatibility, nothing stronger.

The cases outside the envelope are the informative ones. Under Replay, nine are positive and two are negative. So visual assignment can add per-scene value even when selections are concentrated. Under the learned image policies, the result is one-directional: \textbf{all twelve significant Policy and Policy+DR cases lie below the envelope}. For these policies, similarity assigns the selections to the wrong scenes. Holding the frequencies fixed, random assignment would have scored higher.

Why would similarity assign scenes wrongly? The appearance/layout design suggests an answer. Image policies fail under appearance changes. Generic encoders mostly measure appearance. So a similarity score can flag which scenes are hard, without saying which stored policy is safest there. We fixed a test of this in advance: recompute the gap separately on appearance-perturbed scenes and on layout-perturbed scenes, for the twelve significant cases. The negative gap is larger on the appearance side in 9 of 12 cases. The three exceptions are the fresh-scene Door Policy+DR cases, where the smaller layout subsets ($n{=}30$) carry the larger negative gap with wide intervals. The mechanism holds for most cases, not all.

\section{The library changes what the score means}
\label{sec:quality}
The same rule gets different scores in different libraries. This section measures that dependence with three label-informed procedures. All three are measurement tools, not methods: each uses outcome data that a deployed retriever would not have. Estimating experience quality without exhaustive labels remains open.

\textbf{Equalize the library.} We build sub-libraries whose experiences have nearly equal mean success. Labels choose the sub-library; the rule never sees them. At $K{=}10$, equalization uncovers discrimination that unequal libraries hide. Four of five visual rules become positive, with top-pick quality margins near zero; the policy encoder stays negative (Table~\ref{tab:interv}, top). The same rules score 0.12--0.21 lower in the unequal libraries. At $K{=}3$, the equalized advantages are near zero. We do not conclude that equalization removes the signal there: $A$ is unstable when its denominator is small, and we could not confirm the denominator in those sets. The supported statement is narrower. At $K{=}10$, real discrimination survives once library quality is equalized.

\begin{table}[t]
\centering\scriptsize\setlength{\tabcolsep}{3pt}
\caption{Effect of interventions on the advantage $A$. Equalization and deletion use labels; they are diagnostics, not methods.}
\label{tab:interv}
\resizebox{\columnwidth}{!}{\begin{tabular}{lll}
\toprule
intervention & setting & effect on $A$\\
\midrule
equalize library & $K{=}10$ & ResNet $+0.22$, DINOv2 $+0.18$,\\
 & & raw $+0.13$, CLIP $+0.10$, policy enc.\ $-0.09$\\
 & & (unequal libraries: 0.12--0.21 lower)\\
\midrule
re-rank scores & $K{=}50$ & at most 0.08; concentration unchanged\\
(CSLS, mutual prox.) & random $K{=}10$ & mean absolute effect 0.06--0.14\\
\midrule
delete top pick & $K{=}50$ & CLIP $-0.42\rightarrow+0.15$\\
 & & policy enc.\ $-0.57\rightarrow+0.36$\\
\bottomrule
\end{tabular}}
\end{table}

\textbf{Re-rank the scores.} Hubness corrections adjust similarity scores to reduce the dominance of frequent nearest neighbors~\cite{radovanovic2010hubs}. Untuned CSLS~\cite{conneau2018word} and mutual proximity~\cite{schnitzer2012mutual} barely move the results (Table~\ref{tab:interv}, middle). Their effect is smaller than the effect of equalizing the library.

\textbf{Delete the top pick.} Removing one experience, the rule's top pick, flips two rules from negative to positive at $K{=}50$ (Table~\ref{tab:interv}, bottom). Deletion uses labels, so it is not a method. But it locates the failure. The failure lives in the pairing of a concentrated rule with one weak experience. It is not a geometric artifact that re-ranking can fix.

\begin{figure}[t]
\centering
\includegraphics[width=0.82\columnwidth]{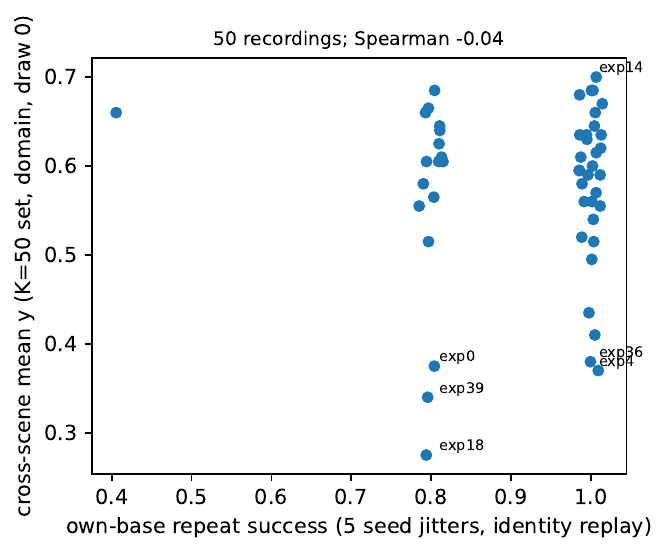}
\caption{Own-base success versus mean cross-scene success for the 50 Replay experiences. The two are uncorrelated ($\rho=-0.04$). Seven experiences have own-base success $\ge0.8$ and cross-scene success $\le0.45$.}
\label{fig:base}
\end{figure}

\textbf{Local success does not imply transfer.} Figure~\ref{fig:base} shows a library problem that exists before any retrieval. Of the 50 Replay experiences, 49 repeat their own base scene with success at least 0.8 (mean 0.92). Yet own-base success does not predict cross-scene success: the correlation is $\rho{=}-0.04$. Seven experiences are reliable at home (own-base $\ge0.8$) and weak elsewhere (cross-scene $\le0.45$). The standard acceptance test for a demonstration asks whether it works where it was collected. That test does not measure what reuse depends on, which is how well the demonstration transfers.

\section{Predicting transfer is not ranking candidates}
\label{sec:pairwise}
The results so far pose a question. If visual selection is weak, does visual distance carry no information? It carries a lot. This section shows where the information goes.

We treat the negated distance $-d(f(q),f(e_i))$ as a score for whether the pair $(e_i,q)$ succeeds, and we measure the area under the ROC curve (AUROC). AUROC is the probability that a random successful pair gets a smaller distance than a random failed pair: 0.5 is chance, 1 is perfect. Pooled over all pairs, the prediction is good for image policies: AUROC reaches 0.85--0.96 under appearance perturbations and 0.70--0.91 under layout perturbations (see also Table~\ref{tab:auroc}). Replay values are lower, as its causal structure requires: appearance cannot affect an open-loop replay.

Pooled AUROC is not selection accuracy. The reason is a decomposition:
\begin{equation}
T(i,q) = \underbrace{g(q)}_{\text{scene difficulty}} + \underbrace{h(i,q)}_{\text{candidate-specific}} + \epsilon .
\end{equation}
A hard scene lowers $T$ for every candidate at once. A score that tracks $g(q)$ predicts many pair outcomes correctly. But selection never compares across scenes. Selection compares the $K$ candidates within one scene, and for that only $h(i,q)$ matters.

So we measure ranking within the scene: for each scene $q$, whether $-d(f(q),f(e_i))$ ranks $T(i,q)$ across the $K$ candidates. For each scene, candidates with $y\ge0.8$ are positives and $y\le0.2$ are negatives; scenes without both are excluded; per-scene AUROC is averaged with scene-clustered intervals. We fixed this rule in advance and recomputed pooled AUROC with the same labels for comparison. The result is consistent (Table~\ref{tab:auroc}): \textbf{within-query AUROC is below pooled AUROC in 62 of 65 eligible cases}. The gap is widest at $K{=}50$: pooled AUROC stays above 0.7, while within-query AUROC is at or near chance for every representation except ResNet. The $K{=}3$ image-policy rows carry a caveat. Only 7--19\% of those scenes are eligible, and with one positive and one negative a per-scene AUROC can only be 0, $\tfrac12$, or 1.

\begin{table}[t]
\centering\scriptsize
\caption{Pooled versus within-query AUROC (ranges over the visual representations). Within-query is lower in 62 of 65 eligible cases.}
\label{tab:auroc}
\begin{tabular}{lcc}
\toprule
setting & pooled & within-query\\
\midrule
Stack Replay, $K{=}10$ & 0.70--0.80 & 0.35--0.71\\
Stack Replay, $K{=}50$ & 0.71--0.76 & 0.45--0.52 (all but ResNet)\\
Image policy, $K{=}3$ & 0.68--0.94 & 0.26--0.77 (7--19\% eligible)\\
\bottomrule
\end{tabular}
\end{table}

A variance decomposition says where the mismatch is worst. Let $g(q)$ be the mean of $T(i,q)$ over the $K$ candidates, and let $R_q^2$ be the share of the total variance of $T$ explained by $g(q)$. Large $R_q^2$ means scene difficulty dominates; small $R_q^2$ means the candidates genuinely differ. We predicted $R_q^2>0.5$ everywhere. The prediction failed in exactly two settings, the Stack Replay $K{=}3$ sets, where $R_q^2$ is 0.42 and 0.45. Those are also the settings with the strongest positive visual-selection results. Everywhere else the prediction held: $R_q^2$ is 0.52--0.57 for Replay at $K\ge10$, and 0.71--0.81 for the image policies, exactly where all twelve significant gaps are negative. So $R_q^2$ varies across settings in step with where visual selection works. Our advance prediction covered only the 0.5 threshold, not this ordering, so we report the ordering as supporting evidence, not a confirmed relationship.

A second check agrees. ResNet is the only rule with a positive gap at every Replay library size (Sec.~\ref{sec:scaling}). It is also the only representation above chance in within-query AUROC at $K{=}50$. CLIP and the policy encoder show the reverse: large advantages in magnitude at $K{=}50$ ($-0.42$, $-0.57$) with within-query AUROC near 0.5. Two different summaries of the matrix single out the same representation.

The design target follows. A retrieval representation does not need to predict which scenes are hard. It needs to order the $K$ candidates within one scene by transferability. Selectors should be trained and evaluated with query-grouped ranking objectives, not only with pooled similarity.

The same distinction recurs beyond retrieval. World foundation models for physical AI~\cite{nvidia2025cosmos} are trained and benchmarked on prediction fidelity, while control requires the conditional analogue: ranking candidate actions or plans by outcome within a single state. Our results do not evaluate such systems, but the audit template transfers directly: pooled predictive skill should not be reported as evidence of decision-relevant ranking until the within-state statistic is measured.

\section{Library size changes the conclusion}\label{sec:scaling}
\begin{figure*}[t]
\centering
\includegraphics[width=0.92\textwidth]{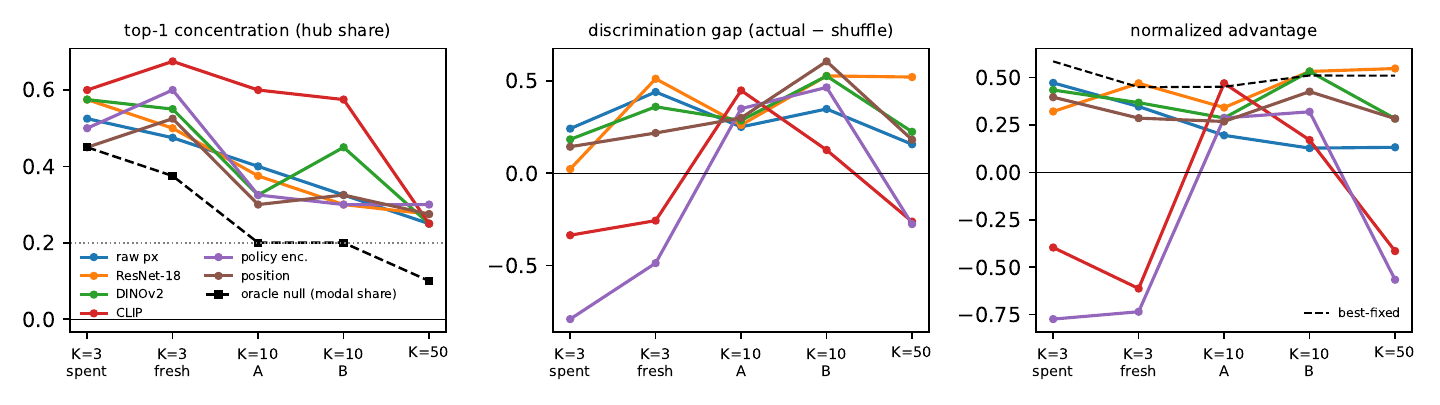}
\caption{Stack Replay at $K{=}3,10,50$: top-pick share against the oracle reference, discrimination gap, and advantage against the best-fixed reference.}
\label{fig:k}
\end{figure*}

The same rule supports opposite conclusions at different library sizes (Fig.~\ref{fig:k}). The policy encoder's gap is $-0.79$ at $K{=}3$, $+0.35$ and $+0.47$ at $K{=}10$, and $-0.28$ at $K{=}50$ (Fig.~\ref{fig:k}, middle). We predicted that the positive $K{=}10$ gaps would persist at $K{=}50$. That held for one rule of four. ResNet is the only rule positive at all three sizes. We report that as a fact about one rule.

Two caveats limit this section. First, the nested libraries extend one recording stream, and the expert's rejection rate changes with size (0.40, 0.33, 0.153 at $K{=}3,10,50$). Size is therefore confounded with recording quality, and the trend is not a scaling law. Random sub-libraries address the concentration part of this concern: excess concentration appears in 96--99\% of 200 random $K{=}10$ compositions, with the identity of the top pick varying widely. Those sub-libraries still share one pool and one query set. Independent recording streams are the stronger test.

Second, the size sensitivity changes how to read our small-library results. Two early conclusions at $K{=}3$ did not survive at $K{=}10$: that recency beat every visual rule, and that one encoder was negatively predictive in general. All image-policy experiments here use $K{=}3$. The learned-policy results are therefore evidence about small libraries, not about scaling.

\section{Two reports every retrieval study can give}
Exhaustive execution is usually infeasible, so the methodology must reduce to cheap reports. The question a reader needs answered is simple. Does the score reflect per-scene selection, or the quality of one repeatedly selected experience? Two reports answer most of it, and any retrieval study can give them.

\textbf{Report the selection distribution.} State how often each experience was selected. This is free: it is a log of the evaluation the study already runs. If exhaustive labels exist, compare against the oracle's distribution, not against uniform. Heavy concentration warns that the score may describe a few experiences.

\textbf{Report best-fixed.} State the success of always using the single best experience, chosen in hindsight. This does not require the full transfer matrix. It requires only the mean success of each library item on the evaluation scenes: $K$ times the rollouts of the standard evaluation, with no per-scene resolution and no repeated seeds. In simulation this is routine. On hardware, a subsample of scenes per item suffices, because identifying the best item needs only rough means. Best-fixed is not a competitor. It shows how much success needed no per-scene selection at all, and how much headroom the retriever can claim.

Table~\ref{tab:survey} shows the current gap. In seven surveyed systems, the two reports never appear together. The two partial cases report the composition of retrieved sources, not per-item selections. This is a reporting gap, not a claim that prior results are wrong. One scope note: the two reports are defined for top-1 selection, while Behavior Retrieval, FlowRetrieval, STRAP, and arguably R+X retrieve a set of data to train on. For set retrieval the analogues are the distribution over retrieved items and training on a single hindsight-chosen subset; the entries for these systems mark reports that would need this translation, not simple omissions. Exhaustive measurement of alternatives has precedent: SRSA~\cite{guo2025srsa} executes every source policy on every prior task to train a transfer predictor, and reports a per-task oracle reference, in a state-based assembly setting with policy-level retrieval. It reports neither the selection distribution nor a best-fixed reference, which is the gap the two reports close.

\begin{table}[t]
\centering
\scriptsize\setlength{\tabcolsep}{2pt}
\caption{Reporting audit of seven retrieval systems. Inclusion criterion: a stored demonstration, trajectory, skill, or training example is selected by an explicit rule before reuse. ``Partial'' means a distribution over source categories, not per-item selections.}
\label{tab:survey}
\resizebox{\columnwidth}{!}{\begin{tabular}{lccl}
\toprule
system & pick dist. & best-fixed & what it does report\\
\midrule
VINN~\cite{pari2022vinn} & no & no & downstream success vs.\ baselines\\
DINOBot~\cite{dipalo2024dinobot} & no & no & per-task success, ablations\\
R+X~\cite{papagiannis2025rx} & no & no & success; varies retrieved-clip count\\
RT-Cache~\cite{kwon2025rtcache} & no & no & success and operation time; varies snippet horizon $N$; latency\\
Behavior Retrieval~\cite{du2023behaviorretrieval} & no & no & relevance separation vs.\ GT labels; GT same-task filter\\
FlowRetrieval~\cite{lin2024flowretrieval} & partial & no & composition: useful / adversarial / non-harmful\\
STRAP~\cite{memmel2025strap} & partial & no & source-task distribution; varies $K$\\
\bottomrule
\end{tabular}}
\end{table}

The two reports also give a way to read any new result. A rule that spreads its selections and beats best-fixed has shown per-scene skill. A rule that concentrates and lands near best-fixed has shown that one experience is good. A rule below its frequency-matched baseline, like the image-policy cases here, is assigning its selections to the wrong scenes.

Where trial history exists, the next step is practical: estimate each experience's cross-scene quality before selecting per scene. Collection-time success is not that estimate (Fig.~\ref{fig:base}). Future work should pair quality estimation with query-conditional ranking, and test both on independent libraries, larger policy libraries, more tasks, and hardware.

\section{Limitations and conclusion}
This is an audit of one benchmark. All $K{=}50$ results use Stack under Replay. All learned-policy results use $K{=}3$, one policy class, and three policies per level. Forty scenes give wide intervals. The frequency-matched envelope has no power analysis, so the 42 of 65 count shows compatibility, not equivalence. The random sub-libraries share one recording pool and one query set. The nested libraries differ in expert rejection rate. All experiments are in simulation. The variance decomposition supports the proposed mechanism but does not confirm it, and the $K{=}3$ within-query values rest on few eligible scenes.

Within this scope, the conclusion is methodological. End-task success does not show whether a retriever chose well. Applying the audit overturned conclusions that ordinary evaluation had suggested, including several of our own. A representation can predict which experience-scene pairs succeed and still fail to rank the candidates for one scene. When library quality is unequal, that failure makes the score follow the rule's top pick. At $K\ge10$, visual rules are 1.5--3 times as concentrated as the oracle, and changing the library moves scores more than re-ranking does. For learned image policies, every significant departure from the frequency-matched baseline is a departure downward.

The recommendation costs little and changes how retrieval results are read. Report which experiences were selected. Report best-fixed. Where counterfactual labels exist, add the frequency-matched and within-query analyses. For future methods, the target is the capability this audit measured and found missing: ranking the candidates within one scene. At $K{=}50$, visual similarity performs that ranking no better than chance (AUROC 0.45--0.52; chance is 0.5) for four of the five embeddings we tested.

\bibliographystyle{IEEEtran}
\bibliography{refs}

@inproceedings{guo2025srsa, title={{SRSA}: Skill Retrieval and Adaptation for Robotic Assembly Tasks}, author={Guo, Yijie and Tang, Bingjie and Akinola, Iretiayo and Fox, Dieter and Gupta, Abhishek and Narang, Yashraj}, booktitle={International Conference on Learning Representations (ICLR)}, year={2025}, note={arXiv:2503.04538}}

@inproceedings{lin2024flowretrieval, title={{FlowRetrieval}: Flow-Guided Data Retrieval for Few-Shot Imitation Learning}, author={Lin, Li-Heng and Cui, Yuchen and Xie, Amber and Hua, Tianyu and Sadigh, Dorsa}, booktitle={Conference on Robot Learning (CoRL)}, year={2024}, note={arXiv:2408.16944}}

@inproceedings{memmel2025strap, title={{STRAP}: Robot Sub-Trajectory Retrieval for Augmented Policy Learning}, author={Memmel, Marius and Berg, Jacob and Chen, Bingqing and Gupta, Abhishek and Francis, Jonathan}, booktitle={International Conference on Learning Representations (ICLR)}, year={2025}, note={arXiv:2412.15182}}

@inproceedings{mandlekar2023mimicgen, title={{MimicGen}: A Data Generation System for Scalable Robot Learning using Human Demonstrations}, author={Mandlekar, Ajay and Nasiriany, Soroush and Wen, Bowen and Akinola, Iretiayo and Narang, Yashraj and Fan, Linxi and Zhu, Yuke and Fox, Dieter}, booktitle={Conference on Robot Learning (CoRL)}, year={2023}, note={arXiv:2310.17596}}

@inproceedings{dipalo2024dinobot, title={{DINOBot}: Robot Manipulation via Retrieval and Alignment with Vision Foundation Models}, author={Di Palo, Norman and Johns, Edward}, booktitle={IEEE International Conference on Robotics and Automation (ICRA)}, year={2024}, note={arXiv:2402.13181}}

@inproceedings{mandlekar2021robomimic, title={What Matters in Learning from Offline Human Demonstrations for Robot Manipulation}, author={Mandlekar, Ajay and Xu, Danfei and Wong, Josiah and Nasiriany, Soroush and Wang, Chen and Kulkarni, Rohun and Fei-Fei, Li and Savarese, Silvio and Zhu, Yuke and Mart{\'i}n-Mart{\'i}n, Roberto}, booktitle={Conference on Robot Learning (CoRL)}, year={2021}}

@article{zhu2020robosuite, title={robosuite: A Modular Simulation Framework and Benchmark for Robot Learning}, author={Zhu, Yuke and Wong, Josiah and Mandlekar, Ajay and Mart{\'i}n-Mart{\'i}n, Roberto and Joshi, Abhishek and Nasiriany, Soroush and Zhu, Yifeng and Lin, Kevin}, journal={arXiv preprint arXiv:2009.12293}, year={2020}}

@article{oquab2024dinov2, title={{DINOv2}: Learning Robust Visual Features without Supervision}, author={Oquab, Maxime and Darcet, Timoth{\'e}e and Moutakanni, Th{\'e}o and others}, journal={Transactions on Machine Learning Research}, year={2024}}

@inproceedings{radford2021clip, title={Learning Transferable Visual Models From Natural Language Supervision}, author={Radford, Alec and Kim, Jong Wook and Hallacy, Chris and others}, booktitle={International Conference on Machine Learning (ICML)}, year={2021}}

@article{radovanovic2010hubs, title={Hubs in Space: Popular Nearest Neighbors in High-Dimensional Data}, author={Radovanovi{\'c}, Milo{\v{s}} and Nanopoulos, Alexandros and Ivanovi{\'c}, Mirjana}, journal={Journal of Machine Learning Research}, volume={11}, pages={2487--2531}, year={2010}}

@inproceedings{pari2022vinn, title={The Surprising Effectiveness of Representation Learning for Visual Imitation}, author={Pari, Jyothish and Shafiullah, Nur Muhammad and Arunachalam, Sridhar Pandian and Pinto, Lerrel}, booktitle={Robotics: Science and Systems (RSS)}, year={2022}, note={arXiv:2112.01511}}

@inproceedings{papagiannis2025rx,
  title={{R+X}: Retrieval and execution from everyday human videos},
  author={Papagiannis, Georgios and Di Palo, Norman and Vitiello, Pietro and Johns, Edward},
  booktitle={IEEE International Conference on Robotics and Automation (ICRA)},
  year={2025},
  note={arXiv:2407.12957}
}

@inproceedings{kwon2025rtcache,
  title={{RT-Cache}: Training-free retrieval for real-time manipulation},
  author={Kwon, Owen and George, Abraham and Bartsch, Alison and Barati Farimani, Amir},
  booktitle={IEEE-RAS International Conference on Humanoid Robots (Humanoids)},
  year={2025},
  note={arXiv:2505.09040}
}

@inproceedings{du2023behaviorretrieval,
  title={Behavior retrieval: Few-shot imitation learning by querying unlabeled datasets},
  author={Du, Maximilian and Nair, Suraj and Sadigh, Dorsa and Finn, Chelsea},
  booktitle={Robotics: Science and Systems (RSS)},
  year={2023},
  note={arXiv:2304.08742}
}

@inproceedings{conneau2018word,
  title={Word Translation Without Parallel Data},
  author={Conneau, Alexis and Lample, Guillaume and Ranzato, Marc'Aurelio and Denoyer, Ludovic and J{\'e}gou, Herv{\'e}},
  booktitle={International Conference on Learning Representations (ICLR)},
  year={2018},
  note={arXiv:1710.04087}
}

@article{schnitzer2012mutual,
  title={Local and Global Scaling Reduce Hubs in Space},
  author={Schnitzer, Dominik and Flexer, Arthur and Schedl, Markus and Widmer, Gerhard},
  journal={Journal of Machine Learning Research},
  volume={13},
  pages={2871--2902},
  year={2012}
}

@article{nvidia2025cosmos,
  title={Cosmos World Foundation Model Platform for Physical {AI}},
  author={{NVIDIA} and Agarwal, Niket and others},
  journal={arXiv preprint arXiv:2501.03575},
  year={2025}
}

\end{document}